\documentclass{bmvc2k}

\usepackage{amsmath}
\usepackage{bm}

\title{Dual Feature Augmentation Network for Generalized Zero-shot Learning}

\addauthor{Lei Xiang}{xl294487391@gmail.com}{1}
\addauthor{Yuan Zhou }{zhouyuan@nuist.edu.cn}{1$\dagger$}
\addauthor{Haoran Duan}{haoran.duan@ieee.org}{2}
\addauthor{Yang Long}{yang.long@ieee.org}{2}

\addinstitution{
 School of Artificial Intelligence\\
 Nanjing University of Information Science and Technology\\
 Nanjing, China
}
\addinstitution{
 Department of Computer Science\\
 Durham University\\
 Durham, UK
}

\runninghead{Lei Xiang, Yuan Zhou, Haoran Duan, Yang Long}{DFAN-GZSL}

\def\etal{\emph{et al}\bmvaOneDot}

\begin{document}

\def\thefootnote{$\dagger$}\footnotetext{Corresponding author}
\maketitle

\begin{abstract}
Zero-shot learning (ZSL) aims to infer novel classes without training samples by transferring knowledge from seen classes. 
Existing embedding-based approaches for ZSL typically employ attention mechanisms to locate attributes on an image.
However, these methods often ignore the complex entanglement among different attributes' visual features in the embedding space. 
Additionally, these methods employ a direct attribute prediction scheme for classification, which does not account for the diversity of attributes in images of the same category.
To address these issues, we propose a novel Dual Feature Augmentation Network (DFAN),  which comprises two feature augmentation modules, one for visual features and the other for semantic features.
The visual feature augmentation module explicitly learns attribute features and employs cosine distance to separate them, thus enhancing attribute representation.
In the semantic feature augmentation module, we propose a bias learner to capture the offset that bridges the gap between actual and predicted attribute values from a dataset's perspective.
Furthermore, we introduce two predictors to reconcile the conflicts between local and global features.
Experimental results on three benchmarks demonstrate the marked advancement of our method compared to state-of-the-art approaches.
Our code is available at \url{https://github.com/Sion1/DFAN}
\end{abstract}

\section{Introduction}
\label{sec:introduction}
The effectiveness of deep learning algorithms in recognizing various image types heavily relies on a significant amount of annotated data.
However, the process of obtaining labeled data for all categories can be a daunting and impractical task. 
To tackle this challenge, ZSL offers a feasible solution by transferring knowledge from seen to unseen classes.
ZSL encompasses two distinct settings depending on the testing approach: conventional ZSL (CZSL)~\cite{lampert2009learning,palatucci2009zero} and generalized ZSL (GZSL)~\cite{chao2016empirical,xian2018zero}. 
CZSL only detects unseen classes, whereas GZSL accounts for both seen and unseen classes during testing. 

Both CZSL and GZSL transfer knowledge from seen classes to unseen classes by semantic information such as manually-annotated attributes~\cite{jayaraman2014zero}, sentence features~\cite{reed2016learning}, and word vectors~\cite{socher2013zero}. 
Among them, attributes as shared class-level semantic features have gained significant popularity.
However, a class's attribute vector is computed by averaging the probability of the attribute occurrence across all images within that class~\cite{welinder2010caltech, xian2017zero, patterson2012sun}, which results in inaccurate predictions due to the diversity of attributes in different images belonging to the same category.
To address this limitation, we propose a bias learner that augments the class-level semantic features for every image.

On the other hand, relying solely on the seen classes during the training process makes models more prone to bias toward these seen classes. 
Therefore, generative-based methods~\cite{verma2018generalized,xian2018feature,schonfeld2019generalized,yue2021counterfactual,chen2021free} are becoming increasingly popular as they complement the information on the unseen classes. 
These approaches generate visual features or images of the unseen classes, thus transforming the ZSL task into a conventional classification problem and significantly reducing the bias towards the seen classes. 
The generative-based methods achieve considerable success but still encounter several challenges.
These challenges encompass optimization difficulties stemming from the notable disparity between synthetic and authentic features, as well as extensive resource consumption during training attributed to the model's substantial size.

In order to avoid dependence on generative models, alternative approaches~\cite{akata2015label,romera2015embarrassingly,chen2018zero,xie2019attentive,xu2020attribute,chen2022transzero, gao2023privacy} learn a single mapping function to fuse different granularity features (\textit{i.e.,} global feature and local feature) and align them with the corresponding semantic features. 
However, the global features are class-oriented while the local features are attribute-oriented, and employing a single mapping function will lead to a decline in the model's performance.
In contrast to these methods, our approach employs two distinct mapping functions (local predictor and global predictor) for multilevel features to achieve visual-to-semantic mapping.

Furthermore, most embedding-based methods employ attributes as a mediator between seen and unseen classes, but these attributes are only represented in a limited number of regions. 
Consequently, utilizing global image features directly is not reasonable, and some methods~\cite{du2022boosting, huynh2020fine, wang2021dual, ge2022dual, chen2022transzero} integrate attention mechanisms~\cite{vaswani2017attention, duan2023dynamic} into the model to locate attribute in the image. 
Despite their effectiveness, they neglect the entanglement of the attribute features in the embedding space.
We explicitly extract attribute features and adopt a cosine similarity loss function to overcome this limitation and enhance visual features.

Our contribution is summarized as follows:
(1) We propose a semantic feature augmentation containing a bias learner, which estimates an offset to alleviate the difference between the actual and predicted attributes, leading to improved class-level semantic features for each image. 
(2) We employ two mapping functions to reinforce the mapping procedure of varying levels of features into the semantic space, thus avoiding inconsistencies in the mapping process of different granular features. 
(3) We propose a visual feature augmentation that explicitly extracts attribute features and adopts a cosine similarity loss to decouple them in the embedding space, thus enhancing the visual features.
(4) We conduct comprehensive experiments on three ZSL benchmarks, demonstrating that our method achieves superior or competitive performance compared to state-of-the-art ZSL methods.
Furthermore, our ablation studies on different modules support the effectiveness of our proposed approach.

\section{Related Works}
\label{sec:related_works}

\subsection{Generalized Zero-shot Learning}
\label{sec:gzsl}
In generalized zero-shot learning (GZSL), modern techniques can be broadly classified into generative-based and embedding-based. 
Among embedding-based approaches, Li \etal~\cite{li2019rethinking} proposes an episode-based training scheme to enhance the generalization ability of semantic embedding. 
Li \etal~\cite{li2018discriminative} introduces an augmented space incorporating class relationships, improving feature discrimination. 
Min \etal~\cite{min2020domain} incorporates a domain detector to select perceived categories, thus reducing the impact of perceived categories on visual-semantic alignment. 
However, these methods depend solely on global features, which poses a challenge in extracting fine-grained information to establish associations between visual and semantic features.

In contrast, generative-based methods in GZSL aim to synthesize visual features of unseen classes by leveraging the shared semantic features. 
Xian \etal~\cite{xian2018feature} proposes a GAN-based approach that employs Wasserstein GAN~\cite{arjovsky2017wasserstein} to generate visual features. 
Schonfeld \etal~\cite{schonfeld2019generalized} utilizes VAE~\cite{kingma2013auto} to align the distributions learned from images and side-information, constructing latent features that incorporate the necessary multi-modal information linked with unseen classes. 
Since these methods generate visual features to complement the information of unseen classes, the issue of unseen samples being misclassified as seen classes is mitigated. 
However, these methods also emphasize learning global image features containing noise, constraining their transferability.

\subsection{Part-based ZSL}
\label{sec:part_based_zsl}
In order to mitigate the impact of noise caused by image-level representations, recent methods have focused on investigating significant regions.
Xu \etal~\cite{xu2020attribute} utilizes a region search approach to identify the regions associated with each attribute after learning a set of attribute prototypes.
Liu \etal~\cite{liu2021goal} introduces a gaze estimation module to determine visual attention regions from the perspective of the human gaze.
Wang \etal~\cite{wang2021dual} progressively adjusts prototypes based on different images and introduces class prototypes to enhance category discriminability.
Ge \etal~\cite{ge2022dual} proposes a method that incorporates all components and yields a binary mask for identifying significant object regions, facilitating the derivation of refined features that serve as the basis for generating critical weights. 
Chen \etal~\cite{chen2022msdn} proposes a dense attribute-based attention mechanism by calculating the similarity between visual and semantic features. 

Although recent methods have shown promising results, they ignore differences in representations of the same attribute due to image variations.
This limitation weakens the representativeness of the features, thus hindering accurate classification.
Moreover, these methods typically employ only one mapping function for learning visual-to-semantic mapping. 
Since there are varying levels of visual features, using the same function leads to conflicts.
We propose using two mapping functions for different visual features to address these issues. 
Additionally, we design a cosine similarity loss that boosts the distance between attribute features.

\section{Method}
\label{sec:method}
\paragraph{Problem Definition.}
This paper focuses on GZSL, which has two disjoint sets.
Let $S=\left\{ x_{i}^{s},y_{i}^{s} \right\} _{i=1}^{N_s}$ denotes the training set (seen classes) consisting of $C_{s}$ classes, where $x_{i}^{s}$ is the $i$-th seen sample and $y_{i}^{s}$ is the corresponding label.
The testing set (unseen classes) $U=\left\{ x_{i}^{u},y_{i}^{u} \right\} _{i=1}^{N_u}$ consists of $C_{u}$ classes, where $x_{i}^{u}$ is the $i$-th unseen sample and $y_{i}^{u}$ is the corresponding label.
Besides, we define the shared semantic features as matrix $A=\left\{ a_j \right\} _{j=1}^{C}\in \mathcal{R} ^{C\times M}$ where $C=C_s+C_u$ is the total number of classes  and $M$ is the number of attributes. 
In matrix $A$, each row $a_j \in \mathcal{R}^M$ represents the semantic feature for $j$-th class.
Each element in vector $a_j$, composed of $M$ attributes, represents the likelihood of the corresponding attribute's occurrence within the $j$-th class.

\begin{figure}[t]%
\centering
\includegraphics[width=1.0\textwidth]{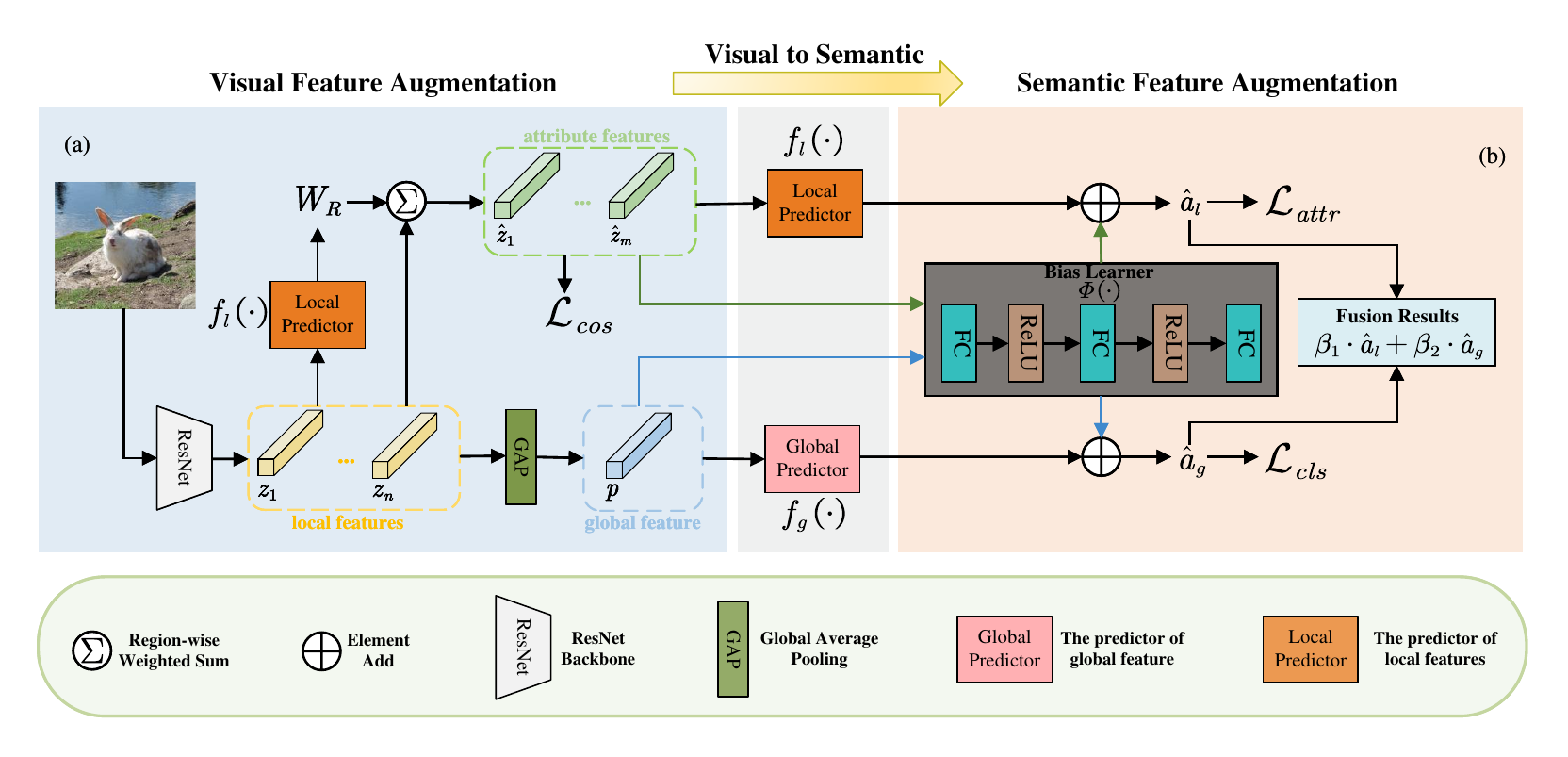}
\caption{
The architecture of DFAN includes two modules:  
(a) The visual feature augmentation module extracts attribute features and employs a cosine similarity loss to disentangle them.
(b) The semantic feature augmentation module aims to learn an offset between actual and predicted values to enhance the semantic features for each image. 
In addition, two predictors are utilized to map different levels of visual features to the semantic space more effectively.
Finally, we fuse different granular feature predictions as the final results to enhance the model's generalization to different datasets.
}
\label{fig:architecture}
\end{figure}

\paragraph{Overview.}
In this paper, we propose a Dual Feature Augmentation Network (DFAN) that leverages two feature augmentation modules to facilitate knowledge transfer from seen to unseen classes. 
Specifically, DFAN consists of two modules, the visual feature augmentation module, and the semantic feature augmentation module, as depicted in Figure~\ref{fig:architecture}.
The visual feature augmentation module extracts the attribute features explicitly by an attention mechanism. 
To enable the mapping of different levels of visual features to the semantic space, two predictors are employed to map visual features of different levels. 
Furthermore, a cosine similarity loss is integrated to disentangle visual features of different attributes.
Concurrently, the semantic feature augmentation module acquires an offset to mitigate the difference between the actual and predicted attribute values.

\subsection{Visual Feature Augmentation}
\label{subsec:attribute_feature_leaner}
In our approach, ResNet~\cite{he2016deep} is used for feature extraction. 
Given an image $x$, the local features $Z=\left\{ z_{i}\right\} _{i=1}^{N=H \times W}$ are extracted from the second last layer of ResNet, where $z_{i}\in \mathcal{R}^{D}$ represents the feature of $i$-th region and $D$ is the embedding dimension.
After that, the global feature $p \in \mathcal{R}^{D}$ is obtained through a global average pooling layer.
To extract attribute features, we follow Ke Zhu and Jianxin Wu~ \cite{zhu2021residual} to adopt an attention mechanism.
Instead of modifying the logit directly, we impose constraints on the attribute features and combine the prediction results using two predictors.
Specifically, we define two distinct linear layers $f_{l}\left( \cdot \right)$ and $f_{g}\left( \cdot \right)$ without bias term as the local predictor and global predictor, respectively.
This architecture enables the efficient mapping of visual features at different levels to the semantic space and avoids potential conflicts between them.

We first use the local predictor to get the local features predict score as matrix $A_{l}=f_{l}(Z)$ where $A_{l}=\left\{ a_{j}^{l} \right\} _{j=1}^{N} \in \mathcal{R}^{N \times M}$. 
In matrix $A_{l}$, each row $a_{j}^{l}$ corresponds to the predicted score for the $j$-th region. 
Specifically, the element $a_{i,j}^{l}$ in the vector $a_{j}^{l}$ represents the probability of attribute $j$ appearing at the $i$-th region.
Subsequently, we obtain the weight of each attribute for each region by applying a softmax operation along the rows, given by:
\begin{equation}
    w_{i,j}^{r} = \frac{\exp (a_{i,j}^{l})}{\sum_{k=1}^{M}\exp(a_{k,j}^{l})}
    \qquad \left( 1\leq i\leq N, 1\leq j\leq M\right),
    \label{eq:W_region}
\end{equation}
where $w_{i,j}^{r}$ is the $i$-th \textbf{r}egion weight for $j$-th attribute.
These weights collectively form a matrix $W_{R} \in \mathcal{R}^{N \times M}$, containing the weights of each region for every attribute.
Subsequently, the $j$-th attribute feature $\hat{z}_{j} \in \mathcal{R}^{D}$ is calculated by region-wise weighted sum as:
\begin{equation}
    \hat{z}_{j} = \varSigma _{k=1}^{N}\,\,w_{k,j}^{r} \cdot z_{k},
    \label{eq:z_attr}
\end{equation}
Similarly, we can get all attribute features as 
$\hat{Z}=\left[ \hat{z}_{1},...,\hat{z}_{M} \right]$.
In order to enhance visual features, we address the entanglement of attributes in the embedding space by reducing their cosine similarities.
Specifically, we compute cosine similarities among attribute features and obtain a similarity matrix $\hat{Z}^{T} \hat{Z}$. 
The diagonal values in this matrix represent the similarities between the feature and itself, while the other values represent the similarity between different attributes. 
To decouple attribute features, we make the diagonal values one and the other zero. 
The ideal similarity matrix should be an identity matrix.
Then, we introduce $\mathcal{L}_{cos}$ loss to minimize the L2 norm between the similarity and identity matrix as follows:
\begin{equation}
    \mathcal{L}_{cos} = \left\| \left. \hat{Z}^{T} \hat{Z} - I \right\| \right. _2,
    \label{eq:L_cos}
\end{equation}
where $I \in \mathcal{R}^{M}$ is the identity matrix.

\subsection{Semantic Feature Augmentation}
\label{subsec:bias_leaner}
To obtain attribute offset for each sample, we design a shared multilayer perceptron $\phi(\cdot)$ to learn such offset.
Specifically,  we use a 3-layer fully connected layer with ReLU as the activation function after the first two layers.
This architecture is chosen because of its capability to capture complex non-linear relationships between input features and attribute offset.
Then, we obtain the final prediction result of attribute features $\hat{Z}$ as:
\begin{equation}
    \hat{a}_{l}= \left< W_{l}, \hat{Z} \right>  + \frac{1}{M}\varSigma _{k=1}^{M}\phi( \hat{z}_{k}),
    \label{eq:hat{a}_l}
\end{equation}
where $W_{l} \in \mathcal{R}^{D \times M}$ is the linear layer weight of the local predictor and $\left< \cdot, \cdot \right>$ denotes the column-wise inner product.
The second term in Eq.~\ref{eq:hat{a}_l} corresponds to an offset between predicted and actual attribute values.
Notably, the inner product of a D-dimensional vector and the columns in $W_{l}$ yield attribute appearance probabilities.
Leveraging the attribute features we have, our task is simplified to compare these features with the matrix. 
The probability of attribute occurrence is consequently obtained through inner product operations on corresponding columns.
Then, the model is updated using the cross-entropy loss as:
\begin{equation}
    \mathcal{L}_{attr} = -\log \frac{\exp \left< \hat{a}_{l}, a_y \right>}{\sum_{k=1}^{C_s}\exp{\left< \hat{a}_{l}, a_{k} \right>}},
    \label{eq:L_attr}
\end{equation}
where $a_y$ denotes the ground truth semantic feature of $x$.

Analogously, we use the same approach to get the prediction result of global feature $p$:
\begin{equation}
    \hat{a}_{g}= f_{g}(p) + \phi(p),
    \label{eq:a_cls}
\end{equation}
it is worth noting that the global predictor $f_{g}(\cdot)$ is used here instead of the local predictor, which can effectively avoid the conflict of different granular features during the mapping process.
Subsequently, a cross-entropy loss is adopted to update the model as:
\begin{equation}
    \mathcal{L}_{cls} = -\log \frac{\exp \left< \hat{a}_{g}, a_y \right>}{\sum_{k=1}^{C_s}\exp{\left< \hat{a}_{g}, a_{k} \right>}}.
    \label{eq:L_cls}
\end{equation}
Finally, we define the overall loss function of our model as follows:
\begin{equation}
    \mathcal{L}_{total} = \mathcal{L}_{attr} + \mathcal{L}_{cls} + \lambda \mathcal{L}_{cos},
    \label{eq:L_total}
\end{equation}
where $\lambda$ is the hyperparameter to control the cosine similarity loss.

\subsection{Zero-shot Prediction}
\label{subsec:zero_shot_prediction}
\paragraph{CZSL prediction.}
After training the model, given an image $x$, we first extract its global feature $p$ and local features $Z$ using ResNet.
Subsequently, we derive attribute features $\hat{Z}$ from the above local features using the attention mechanism.
After that, the prediction results $\hat{a}_{g}$ and $\hat{a}_{l}$ are obtained by two predictors, respectively.
Finally, we combine these scores, controlled by coefficients $(\beta1, \beta2)$, to get a prediction score and predict the category of the input image by:
\begin{equation}
    \hat{y} = \underset{j\in C_{u}}{arg\max} \quad a_{j}^{T}(\beta1 \cdot \hat{a}_{l} + \beta2 \cdot \hat{a}_{g})  
    \label{eq:y_czsl}
\end{equation}

\paragraph{GZSL prediction.}
For GZSL, we need to consider both seen and unseen classes. 
We introduce Calibrated Stacking (CS)~\cite{chao2016empirical} to reduce the prediction bias towards seen classes.
The prediction category is:
\begin{equation}
    \hat{y} = \underset{j\in C_{s}\cup C_{u}}{arg\max} \quad a_{j}^{T} (\beta1 \cdot \hat{a}_{l} + \beta2 \cdot \hat{a}_{g}) 
    - \gamma \cdot \mathcal{I} (x\in S)
    \label{eq:y_gzsl}
\end{equation}
where the indicator $\mathcal{I}(\cdot) \in \left\{0, 1\right\}$ indicates whether or not $x$ is a seen class and $\gamma$ is a calibration factor.

\section{Experiments and Analysis}
\label{sec:experiment}

\subsection{Experimental Settings}
\label{sec:experiment_setting}
\paragraph{Dataset.}
We evaluated our proposed method on three benchmark datasets for GZSL: Animals with Attributes 2 (AWA2\cite{xian2018zero}), Caltech-UCSD Birds-200-2011 (CUB)\cite{wah2011caltech}, and SUN Attribute (SUN)\cite{patterson2012sun}.
AWA2 consists of 37,322 images of 85 different animal species, CUB contains 11,788 images of 200 bird species, and SUN comprises 14,340 images of 717 scenes, each class annotated with 102 attributes.
In order to serve as semantic descriptors, we utilized the pre-defined attributes for each dataset. Additionally, we adopted the Proposed Split\cite{xian2018zero} to divide all classes into seen and unseen classes on each dataset, which allowed for a more comprehensive evaluation of our method's effectiveness in handling the GZSL problem.

\paragraph{Evaluation.}
The evaluation methodology proposed in \cite{xian2018zero} is adopted in our work.
For the CZSL scenario, we only measure the per-class Top-1 accuracy on the unseen classes.
For the GZSL scenario, we evaluate the model's performance on both seen and unseen classes, represented as $S$ and $U$, respectively.
To provide a comprehensive evaluation of the model's GZSL performance, we use the harmonic mean, denoted as $H$, which is a suitable performance metric and calculated as $H= {(2\times S\times U)}/{(S+U)}$.

\paragraph{Implement Details.}
We use ResNet101 as the backbone of our model, which is pre-trained on ImageNet1k. 
It is worth noting that the classes in ImageNet1k do not overlap with the unseen classes in the benchmark datasets. 
The Adam optimizer is utilized for finetuning the ResNet with a learning rate of 0.00001 and weight decay of 0.0001.
We use a learning rate of 0.001 and weight decay of 0.00001 to optimize the linear layer.  
All experiments are conducted on an NVIDIA RTX 3090 GPU, Intel processor, a memory-sized 32GB, and trained with 80 epochs.
Regarding the combined coefficients, we set (0.5, 0.5) for CUB and SUN and (0.0, 1.0) for AWA2 since class-level features are more crucial in coarse-grained datasets like AWA2. 
We implement all our methods using the PyTorch framework.

\subsection{Comparison with State-of-the-Art Methods}
\label{sec:comparison_with_sotas}
\paragraph{Conventional Zero-Shot Learning.}
We compare the performance of our proposed method with the state-of-the-art methods in the CZSL setting.
Table~\ref{Table:compare_with_sotas} presents the results of CZSL on various datasets.
Our proposed method achieves the third-best accuracy of 77.3\% and best accuracy of 67.9\% on CUB and SUN, respectively. 
We are only 0.5\% away from the best performance on CUB and 0.3\% from the second-best performance on this dataset. 
Our method differs from the second-best result on the SUN dataset by 2.1\%. 
These results prove that our method learns better attribute features and achieves superior results on fine-grained datasets. 
However, compared to other non-generative methods, we do not obtain better results on coarse-grained datasets such as AWA2.
The primary reason is AWA2's attribute descriptions, which focus on global image content rather than local details.
This poses a challenge to part-based methods.
Furthermore, unlike other part-based methods like MSDN~\cite{chen2022msdn} and TransZero~\cite{chen2022transzero}, which leverage additional semantic features such as word2vec\cite{mikolov2013efficient} or GloVe\cite{pennington2014glove} for locating attribute, our DFAN only uses the attribute description given by the dataset.

\paragraph{Generalized Zero-Shot Learning.}
Table~\ref{Table:compare_with_sotas} displays the results of different methods in the GZSL setting.
Our method attained the best performance on CUB and the second-best performance on both SUN and AWA2 when compared to other approaches.
In the AWA2 dataset, we only differed from the second-best approach by a margin of 0.1\%.
On the other hand, the generative methods outperform non-generative methods on the SUN dataset.
The reason is that SUN has a large number of categories and uneven image distribution, making it difficult for non-generative methods to learn a robust semantic space.
Conversely, these generative methods effectively enhance SUN performance by generating images or visual features to complement information.
In contrast, our DFAN achieves the best results in the non-generative approaches.
These results suggest that our proposed method delivers superior performance in GZSL and CZSL settings.

\begin{table*}[t]
\resizebox{1.0\linewidth}{!}{
\begin{tabular}{r|ccc|c|ccc|c|ccc|c}
\hline
{\textbf{Methods}} 
		&\multicolumn{4}{c|}{\textbf{CUB}}&\multicolumn{4}{c|}{\textbf{SUN}}&\multicolumn{4}{c}{\textbf{AWA2}}\\
		\cline{2-5}\cline{6-9}\cline{9-13}
		&\multicolumn{3}{c|}{GZSL}&\multicolumn{1}{c|}{CZSL}&\multicolumn{3}{c|}{GZSL}&\multicolumn{1}{c|}{CZSL}&\multicolumn{3}{c}{GZSL}&\multicolumn{1}{c}{CZSL}\\
		\cline{2-5}\cline{6-9}\cline{9-13}
		\textbf{} 
		&\rm{U} & \rm{S} & \rm{H} &\rm{acc}&\rm{U} & \rm{S} & \rm{H} &\rm{acc}&\rm{U} & \rm{S} & \rm{H} &\rm{acc}\\
				
		\hline
		\textbf{Generative Methods}\\ 
		f-VAEGAN-D2~\cite{xian2019f} & 48.4  & 60.1  & 53.6  & 61.0  & 45.1  & \textcolor[rgb]{ 0,  .69,  .941}{\textbf{38.0}} & 41.3  & 64.7  & 57.6  & 70.6  & 63.5  & \textcolor[rgb]{ .329,  .51,  .208}{\textbf{71.1}} \\
    E-PGN~\cite{yu2020episode} & 52.0  & 61.1  & 56.2  & 72.4  &\textbf{–} &\textbf{–} &\textbf{–} &\textbf{–} & 52.6  & \textcolor[rgb]{ 0,  .69,  .941}{\textbf{83.5}} & 64.6  & \textcolor[rgb]{ 1,  0,  0}{\textbf{73.4}} \\
    Composer~\cite{huynh2020compositional} & 56.4  & 63.8  & 59.9  & 69.4  & \textcolor[rgb]{ 1,  0,  0}{\textbf{55.1}} & 22.0  & 31.4  & 62.6  & \textcolor[rgb]{ .329,  .51,  .208}{\textbf{62.1}} & 77.3  & 68.8  & \textcolor[rgb]{ 0,  .69,  .941}{\textbf{71.5}} \\
    GCM-CF~\cite{yue2021counterfactual} & 61.0  & 59.7  & 60.3  &\textbf{–} & 47.9  & \textcolor[rgb]{ .329,  .51,  .208}{\textbf{37.8}} & \textcolor[rgb]{ .329,  .51,  .208}{\textbf{42.2}} &\textbf{–} & 60.4  & 75.1  & 67.0  & \textbf{–} \\
    CE-GZSL~\cite{han2021contrastive} & 63.9  & 66.8  & 65.3  & \textcolor[rgb]{ 0,  .69,  .941}{\textbf{77.5}} & 48.8  & \textcolor[rgb]{ 1,  0,  0}{\textbf{38.6}} & \textcolor[rgb]{ 1,  0,  0}{\textbf{43.1}} & 63.3  & \textcolor[rgb]{ 0,  .69,  .941}{\textbf{63.1}} & 78.6  & 70.0  & 70.4  \\
    FREE~\cite{chen2021free} & 55.7  & 59.9  & 57.7  &\textbf{–} & 47.4  & 37.2  & 41.7  &\textbf{–} & 60.4  & 75.4  & 67.1  & \textbf{–} \\
		\hline 
		\textbf{Non-generative Methods}\\   
		DAZLE~\cite{huynh2020fine} & 56.7  & 59.6  & 58.1  & \textbf{–} & \textcolor[rgb]{ .329,  .51,  .208}{\textbf{52.3}} & 24.3  & 33.2  &\textbf{–} & 60.3  & 75.3  & 67.1  & \textbf{–} \\
    APN~\cite{xu2020attribute} & \textcolor[rgb]{ .329,  .51,  .208}{\textbf{65.3}} & 69.3  & 67.2  & 72.0  & 41.9  & 34.0  & 37.6  & 61.6  & 57.1  & 72.4  & 63.9  & 68.4  \\
    GEM-ZSL~\cite{liu2021goal} & 64.8  & \textcolor[rgb]{ 0,  .69,  .941}{\textbf{77.1}} & \textcolor[rgb]{ 0,  .69,  .941}{\textbf{70.4}} & \textcolor[rgb]{ 1,  0,  0}{\textbf{77.8}} & 38.1  & 35.7  & 36.9  & 62.8  & \textcolor[rgb]{ 1,  0,  0}{\textbf{64.8}} & 77.5  & \textcolor[rgb]{ 1,  0,  0}{\textbf{70.6}} & 67.3  \\
    SR2E~\cite{ge2021semantic} & 61.6  & \textcolor[rgb]{ .329,  .51,  .208}{\textbf{70.6}} & 65.8  &\textbf{–} & 43.1  & 36.8  & 39.7  &\textbf{–} & 58.0  & 80.7  & 67.5  & \textbf{–} \\
    MSDN~\cite{chen2022msdn} & 65.3  & 69.3  & 67.2  & 72.0  & 52.2  & 34.2  & 41.3  & \textcolor[rgb]{ 0,  .69,  .941}{\textbf{65.8}} & 62.0  & 74.5  & 67.7  & 70.1  \\
    TransZero~\cite{chen2022transzero} & \textcolor[rgb]{ 1,  0,  0}{\textbf{69.3}} & 68.3  & \textcolor[rgb]{ .329,  .51,  .208}{\textbf{68.8}} & 76.8  & \textcolor[rgb]{ 0,  .69,  .941}{\textbf{52.6}} & 33.4  & 40.8  & \textcolor[rgb]{ .329,  .51,  .208}{\textbf{65.6}} & 61.3  & \textcolor[rgb]{ .329,  .51,  .208}{\textbf{82.3}}  & \textcolor[rgb]{ .329,  .51,  .208}{\textbf{70.2}} & 70.1  \\
    ours & \textcolor[rgb]{ 0,  .69,  .941}{\textbf{65.4}} & \textcolor[rgb]{ 1,  0,  0}{\textbf{79.7}} & \textcolor[rgb]{ 1,  0,  0}{\textbf{71.8}} & \textcolor[rgb]{ .329,  .51,  .208}{\textbf{77.3}} & 51.0  & 36.4  & \textcolor[rgb]{ 0,  .69,  .941}{\textbf{42.5}} & \textcolor[rgb]{ 1,  0,  0}{\textbf{67.9}} & 58.9  & \textcolor[rgb]{ 1,  0,  0}{\textbf{88.0}} & \textcolor[rgb]{ 0,  .69,  .941}{\textbf{70.5}} & 67.4  \\
				\hline	
		\end{tabular}	
}
  \centering  
		\caption{Results ~($\%$) of the state-of-the-art CZSL and GZSL modes on CUB, SUN and AWA2, including generative methods and non-generative methods. The best, second-best and third-best results are marked in \textcolor[rgb]{ 1,  0,  0}{\textbf{Red}}, \textcolor[rgb]{ 0,  .69,  .941}{\textbf{Blue}} and \textcolor[rgb]{ .329,  .51,  .208}{\textbf{Green}}, respectively. The symbol “\textbf{–}” indicates no results.}\label{Table:compare_with_sotas}
\end{table*}

\begin{table*}[t]
\resizebox{1.0\linewidth}{!}{
\begin{tabular}{c|ccc|c|ccc|c|ccc|c}
\hline
{\textbf{Methods}} 
		&\multicolumn{4}{c|}{\textbf{CUB}}&\multicolumn{4}{c|}{\textbf{SUN}}&\multicolumn{4}{c}{\textbf{AWA2}}\\
		\cline{2-5}\cline{6-9}\cline{9-13}
		&\multicolumn{3}{c|}{GZSL}&\multicolumn{1}{c|}{CZSL}&\multicolumn{3}{c|}{GZSL}&\multicolumn{1}{c|}{CZSL}&\multicolumn{3}{c|}{GZSL}&\multicolumn{1}{c}{CZSL}\\
		\cline{2-5}\cline{6-9}\cline{9-13}
		\textbf{} 
		&\rm{U} & \rm{S} & \rm{H} &\rm{acc}&\rm{U} & \rm{S} & \rm{H} &\rm{acc}&\rm{U} & \rm{S} & \rm{H} &\rm{acc}\\
  \hline
		single predictor & 54.3  & 72.2  & 62.0  & 65.0  & \textbf{43.3}  & 22.8  & 29.8  & 53.6  & \textbf{56.8}  & 76.4  & 65.2  & 60.2  \\
        two predictors & 60.8  & 80.3  & 69.2  & 76.0  & 39.7  & 32.2  & 35.6  & 58.5  & 55.5  & 87.3  & 67.8  & 62.7  \\
        two predictors + bias leaner & \textbf{63.2}  & \textbf{81.2}  & \textbf{71.1}  & \textbf{76.7}  & 38.8  & \textbf{33.2}  & \textbf{35.7}  & \textbf{59.3}  & 55.1  & \textbf{88.9}  & \textbf{68.0}  & \textbf{63.4}  \\
  \hline	
		\end{tabular}	
}
  \centering  
		\caption{Ablation study of different modules. 'single predictor' uses a shared predictor, and 'two predictors' uses two different predictors for different level features.}\label{Table:ablation_study_modules}
\end{table*}

\begin{table*}[t]
\resizebox{1.0\linewidth}{!} {
  \begin{tabular}{c|ccc|c|ccc|c|ccc|c}
\hline
{\textbf{Methods}} 
		&\multicolumn{4}{c|}{\textbf{CUB}}&\multicolumn{4}{c|}{\textbf{SUN}}&\multicolumn{4}{c}{\textbf{AWA2}}\\
		\cline{2-5}\cline{6-9}\cline{9-13}
		&\multicolumn{3}{c|}{GZSL}&\multicolumn{1}{c|}{CZSL}&\multicolumn{3}{c|}{GZSL}&\multicolumn{1}{c|}{CZSL}&\multicolumn{3}{c|}{GZSL}&\multicolumn{1}{c}{CZSL}\\
		\cline{2-5}\cline{6-9}\cline{9-13}
		\textbf{} 
		&\rm{U} & \rm{S} & \rm{H} &\rm{acc}&\rm{U} & \rm{S} & \rm{H} &\rm{acc}&\rm{U} & \rm{S} & \rm{H} &\rm{acc}\\
  \hline
  $\mathcal{L}_{cls}$ & 54.3  & 72.2  & 62.0  & 65.0  & 43.3  & 22.8  & 29.8  & 53.6  & 56.8  & 76.4  & 65.2  & 60.2  \\
  $\mathcal{L}_{attr}$ & 51.8 & 69.4 & 59.3 & 61.5 & 45.3 & 26.5 & 33.5 & 62.0 & 50.6 & 88.4 & 64.4 & 63.7 \\
  $\mathcal{L}_{cls}$+$\mathcal{L}_{attr}$ & 60.8  & \textbf{80.3}  & 69.2  & 76.0  & 39.7  & 32.2  & 35.6  & 58.5  & 55.5  & 87.3  & 67.8  & 62.7  \\
  $\mathcal{L}_{cls}$+$\mathcal{L}_{attr}$+$\mathcal{L}_{cos}$ & \textbf{65.4} & 79.7 & \textbf{71.8} & \textbf{77.3} & \textbf{51.0} & \textbf{36.4} & \textbf{42.5} & \textbf{67.9} & \textbf{58.9} & \textbf{88.0} & \textbf{70.5} & \textbf{67.4} \\
  \hline
\end{tabular}
}
\caption{Ablation study of loss terms in Eq.~\ref{eq:L_total}. We accumulate each term from top to bottom of the table.}
\label{Table:ablation_study_losses}
\end{table*}

\subsection{Ablation Study}
\label{sec:ablation_study_modules}
\paragraph{Effect of Two predictors.}
DFAN employs two different predictors to avoid conflicts between features of different levels. 
Ablation studies are conducted to validate their effectiveness, and the results are presented in Table~\ref{Table:ablation_study_modules}.
Our findings show that using two predictors significantly improves recognition accuracy, particularly in CZSL setting on the CUB dataset, where our method outperforms the single predictor approach by up to 11\%.
Furthermore, in the GZSL setting, compared to utilizing a single predictor, employing two predictors achieves a significant gain of 10\% on average in seen class accuracy while maintaining unseen class accuracy.

\paragraph{Effect of Bias Learner.}
The bias learner effectively corrects attribute prediction differences by learning offsets similar to residual connections. 
Its performance is evaluated in Table~\ref{Table:ablation_study_modules}.
In the CZSL setting, the bias learner improves seen class accuracy by an average of 1\% compared to using two predictors.
In the GZSL setting, our module improves harmonic mean by 1.9\% on CUB, where the dataset is exclusively composed of bird species, and the distinction between classes relies mainly on subtle differences between attributes.
In contrast, we did not achieve significant improvements on the AWA2 and SUN datasets due to their diverse categories and irregular attribute distributions, which makes it difficult for fully connected layers to accurately capture the relationship between attributes.

\begin{figure}[t]%
\centering
\includegraphics[width=1.0\textwidth]{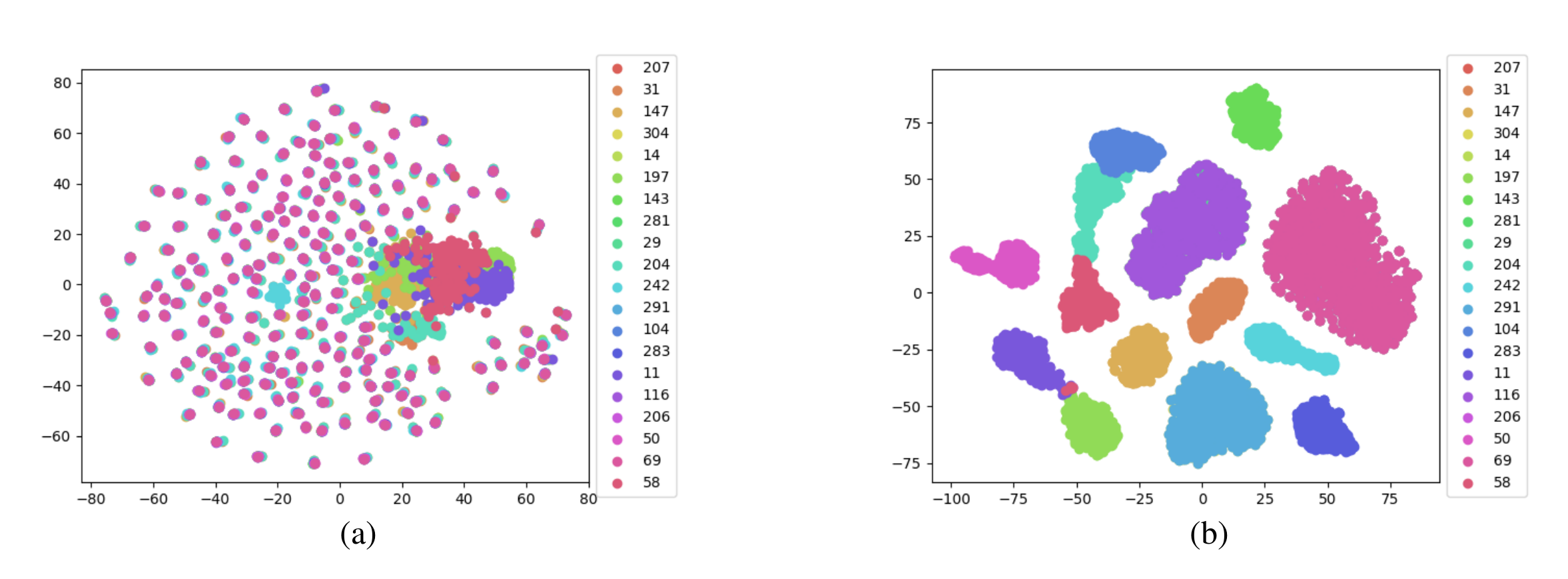}
\caption{T-SNE visualizations of attribute features and the numbers to the right of each legend represent randomly selected attributes. (a) $\mathcal{L}_{cos}$ is not used. (b) $\mathcal{L}_{cos}$ is used.}
\label{fig:t_sne}
\end{figure}

\paragraph{Effect of Loss Function.}
In order to investigate the effectiveness of each of our loss functions for attribute representation learning, we conduct an ablation study by adding losses one by one, starting from the baseline where no attribute features are included in the framework, denoted as $\mathcal{L}_{cls}$. 
The results are presented in Table~\ref{Table:ablation_study_losses}, where it is evident that incorporating $\mathcal{L}_{attr}$ and $\mathcal{L}_{cos}$ both improve the performance and their effects are complementary.
Notably, the cosine similarity loss $\mathcal{L}_{cos}$ has the greatest impact on the overall performance improvement.

\paragraph{t-SNE Visualizations.}
In order to assess the representativeness of our proposed attribute features, we conducted a t-SNE visualization analysis on 20 randomly selected attribute features extracted from multiple images, with and without $\mathcal{L}_{cos}$.
As depicted in Figure~\ref{fig:t_sne}, the attribute features learned with $\mathcal{L}_{cos}$ are distinctly clustered based on the corresponding attributes and exhibit a clear boundary, whereas those without $\mathcal{L}_{cos}$ are relatively mixed.
This observation suggests that incorporating $\mathcal{L}_{cos}$ in our DFAN effectively enhances the discrimination ability of attribute-level features, thereby promoting effective attribute representation learning.

\begin{figure}[t]%
\centering
\includegraphics[width=1.0\textwidth]{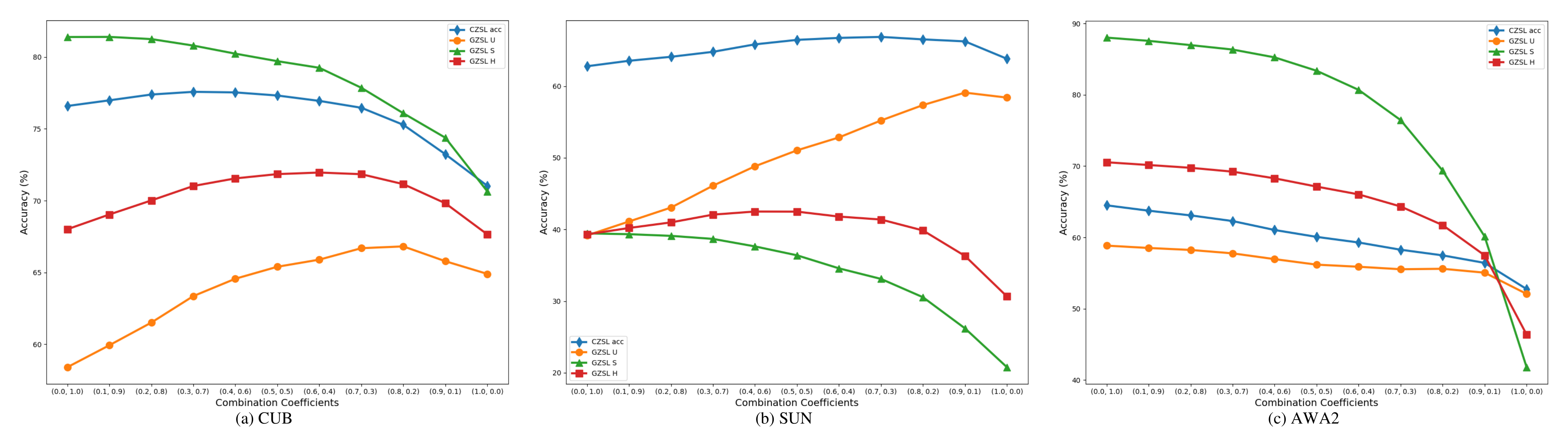}
\caption{Effect of coefficients ($\beta$1, $\beta$2) in the CZSL and GZSL setting.
}
\label{fig:combine_coefficients}
\end{figure}

\subsection{Discussion on $\beta1$ and $\beta2$}
We evaluate the impact of the ratio $\beta$ on the final prediction results, which controls the relative contribution of attribute-level features and class-level features in the proposed model. 
The results are presented in Figure~\ref{fig:combine_coefficients}.
In fine-grained datasets, such as CUB and SUN, the CZSL accuracy and the harmonic mean increase first and then decrease as the attribute features play a more dominant role in these datasets.
Our DFAN performs best with coefficients (0.3,0.7) on CUB and (0.7,0.3) on SUN in CZSL. 
In GZSL, our method's peak performance is at (0.6, 0.4) for CUB, while SUN reaches its best result at (0.5, 0.5).
In contrast, for the coarse-grained dataset AWA2, as the value of $\beta_1$ gradually increases, the CZSL accuracy and the harmonic mean decrease and reach a minimum when $\beta_1$ equals 1.
Our DFAN performs best in CZSL and GZSL when the coefficient is (0.0, 1.0).
This outcome can be attributed to the coarse-grained dataset's attribute descriptions, which encompass the entirety of the image, leading to the dominant influence of class-level features on the ultimate prediction.

\section{Conclusion}
\label{sec:conclusion}
In conclusion, our proposed DFAN effectively addresses the limitations of existing methods in GZSL.
The visual feature augmentation module enhances attribute feature extraction and disentangles them through a cosine similarity loss. 
This module enables better capturing of attribute-specific visual features. 
With the aid of a bias learner, the semantic feature augmentation module effectively handles discrepancies between predicted and actual attribute values, leading to improved shared semantic features for each sample.
Moreover, DFAN utilizes two mapping functions to bridge the visual and semantic features, facilitating effective fusion and linkage of information at different levels.
The results validate the effectiveness of DFAN's visual and semantic feature augmentation modules and the utilization of two mapping functions.

\section*{Acknowledgement}
\label{sec:acknowledgement}
This project is supported by International Exchanges 2022 IEC\textbackslash NSFC\textbackslash223523 and Securing the Energy/Transport Interface EP/X037401/1.

\bibliography{egbib}
\end{document}